\documentclass[sigconf, nonacm]{acmart}
\usepackage{pvldb}
\usepackage{array}
\usepackage{multirow}
\usepackage{amsmath}
\usepackage{placeins}
\usepackage{algorithm}
\usepackage{algpseudocode}
\title{VL-Dedup: Graded Multimodal Entity Resolution for Image-Text Data Deduplication}
\author{Zhenxi Liu}
\affiliation{%
  \institution{Zhejiang University}
  \country{China}
}

\author{Guanze Chen}
\affiliation{%
  \institution{The Chinese University of Hong Kong}
  \country{China}
}

\author{Dehui Yu}
\affiliation{%
  \institution{Zhejiang University}
  \country{China}
}

\author{Xingyu Wang}
\affiliation{%
  \institution{University of Science and Technology Liaoning}
  \country{China}
}

\author{Bingxu Hou}
\affiliation{%
  \institution{Zhejiang University}
  \country{China}
}

\author{Chengwei Li}
\affiliation{%
  \department{Institute of Automation}
  \institution{Chinese Academy of Sciences}
  \country{China}
}

\author{Tao Ma}
\affiliation{%
  \department{Institute of Computing Technology}
  \institution{Chinese Academy of Sciences}
  \country{China}
}

\author{Ke Wang}
\affiliation{%
  \institution{Zhejiang University}
  \country{China}
}

\author{Gongsheng Yuan}
\authornote{Corresponding author.}
\affiliation{%
  \institution{Zhejiang University}
  \country{China}
}
\email{ygs@zju.cn}

\renewcommand{\shortauthors}{Liu et al.}

\begin{document}
\begin{abstract}
Image--text corpora contain copies of one image, captions that describe one
image in different ways, captions reused for different images, and captions
that do not match their images. Each case calls for a different curation
action. Deduplication and multimodal entity resolution (ER) only decide
whether records are the same, and they are evaluated on labeled pairs. Yet
pair scores do not tell which records end up in the same cluster or which
useful captions get deleted. We therefore formulate image--text deduplication
as graded multimodal ER, which decides visual identity, caption handling, and
mismatches for each pair, and we evaluate it at three levels: pairs, clusters,
and the retained set. Our method, VL-Dedup, groups records of the same visual
entity into clusters and caps the size of each cluster. Within each cluster, a
vision--language model (VLM) and CLIPScore decide which captions to keep and
which to flag. We also build VL-DedupBench, with 50,000 COCO and SOP records
and 66,788 labeled pairs. On it, a classifier that selects no false edge among the labeled pairs still joins
millions of record pairs from different entities.
Spending VLM calls on the pairs inside clusters, rather than on pairs with
uncertain scores, gives better caption decisions from the same model. Even
when almost all clusters are correct, keeping one record per cluster loses
facts in many deletions, while the deletions of VL-Dedup lose much less
information.
\end{abstract}

\maketitle
\pagestyle{plain}

\section{Introduction}
\label{sec:introduction}

\begin{figure}[!t]
\centering
\includegraphics[width=\columnwidth]{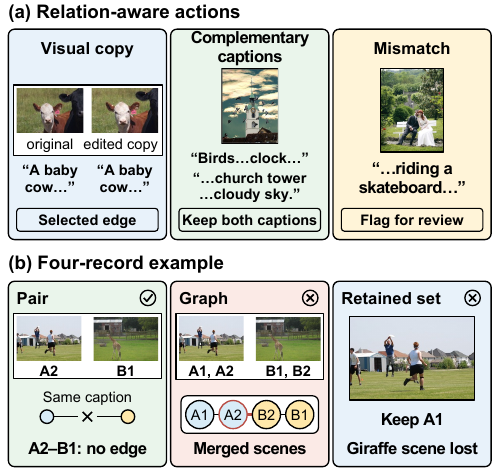}
\caption{(a) Three relations in an image--text corpus and the action that
each needs. (b) Four test records at the pair, graph, and retained-set
levels.}
\label{fig:teaser}
\end{figure}

Large image--text corpora support vision--language pretraining and retrieval.
Their quality depends in part on how corpus construction handles redundant or
contradictory records
\citep{schuhmann2022laion5b,gadre2023datacomp,awadalla2024mint1t}. Consider
the records in Figure~\ref{fig:teaser}(a). A corpus contains copies of one
image, captions that add different facts about one image, captions reused
across different images, and captions that do not correctly describe their
images (mismatched captions). These relations call for different actions.
Copies of one visual entity can be deduplicated, captions with complementary
facts should be kept, and mismatched captions should be flagged for review.
A binary deduplication policy assigns the same action to all of them, so it
can join records of different entities or discard useful captions.

Two lines of work address parts of this problem. Training data deduplication
removes near-duplicate and semantically redundant records
\citep{lee2022deduplm,abbas2023semdedup}. It defines redundancy by surface or
embedding similarity, which does not tell whether two records show the same
visual entity. It is usually evaluated by removal rates or downstream
accuracy. Entity resolution (ER) decides whether records refer to the same
real-world entity using textual, visual, or fused evidence
\citep{fellegi1969record,elmagarmid2007duplicate,li2020ditto,wilke2021multimodaler}.
ER returns one identity decision per pair or one cluster per entity, which
does not say which captions to keep, and ER matchers are evaluated on labeled
pairs.

A curation system makes its decisions in three steps. It decides pairs, it
groups records into clusters along the selected pairs, and it exports a
retained set of records. The user of the corpus sees only the retained set.
ER research has shown that clustering pair decisions changes the result
\citep{hassanzadeh2009clustering,menestrina2010evaluating}. Correct pair
decisions do not ensure correct clusters, and correct clusters do not ensure
a retained set that keeps every fact, so each level needs its own evaluation.
We therefore formulate image--text deduplication as \emph{graded multimodal
ER}. Each pair receives a visual identity decision, a caption action, and a
mismatch action, and we evaluate the decisions on three levels: pairs,
clusters, and the retained set.

We present VL-Dedup, a system for graded multimodal ER that separates visual
identity from caption and mismatch actions. A pair classifier scores
candidate pairs from visual and textual channels and selects edges. VL-Dedup
then adds three steps. (1) It groups records along the selected
edges into visual-entity clusters and limits each cluster to at most $K$
records, with $K$ chosen on validation data. (2) It combines the answer of a VLM about each
record with CLIPScore \citep{hessel2021clipscore} into one record score, which sets mismatch actions.
(3) It asks the VLM for a verdict on the remaining pairs inside each cluster.
The verdicts set caption actions and do not change the clusters, so records
with complementary captions can share a cluster and still keep both
captions. An export policy turns all actions into a
retained set. Figure~\ref{fig:pipeline} shows the six stages of VL-Dedup. We
also build VL-DedupBench, a benchmark of COCO and SOP records on which all
three levels can be scored.

Figure~\ref{fig:teaser}(b) shows four test records of VL-DedupBench. The
letter names the image, and the number names the record. A1 shows people
playing frisbee. A2 has the same image, but its caption, about a giraffe and a
zebra in a grassy field, comes from another image. B1 and B2 show that
giraffe, and B1 has the same caption as A2. A gradient boosting classifier
correctly selects no edge for the labeled pair A2--B1. It also selects the
unlabeled pairs A1--A2, whose records share an image, and A2--B2, whose captions
differ in one word. The selected edges are correct on every labeled pair, yet
the four records share a cluster, and further edges grow this cluster to
33,362 records. If the cluster held only these four records, an export that
keeps the record with the highest CLIPScore would keep A1 and delete B1 and
B2, so the giraffe scene would be lost.

The example shows why VL-Dedup handles each level with its own decision. No
selected edge joins A1 and B1, yet they share a cluster through the path
A1--A2--B2--B1. One wrong edge, A2--B2, joins the two scenes, and every later
edge into either side extends the cluster. Over the whole test partition,
most false pairs of this classifier are joined by several such paths, so
removing single edges cannot separate them. The size cap limits all paths at
once, because no cluster can grow beyond $K$ records.
The record score targets records such as A2, which links the two scenes
because its caption describes another image. Inside each cluster, the VLM decides which
captions add facts, so the export keeps the records with complementary
captions instead of one record per cluster.

In summary, we make the following contributions.

\begin{itemize}
\item We formulate image--text deduplication as graded multimodal ER and
define three evaluation levels: selected edges on labeled pairs, clusters over
all record pairs, and the coverage of deleted records by the retained set
(Section~\ref{sec:problem}).
\item We present VL-Dedup, which separates visual identity from caption and
mismatch actions. It calls the VLM only inside clusters, where an export
decides what to delete, and its pair verdicts set caption actions without
changing the clusters. It flags mismatched captions with a record score that
puts CLIPScore and the probability of the VLM on one scale. With the size
cap, $n$ records need at most $n(K-1)/2$ pair verdicts
(Section~\ref{sec:method}).
\item We build VL-DedupBench, with 50,000 COCO and SOP records, 66,788
labeled pairs, three edited versions of the test images, several photos of
each test product, and larger entities. We release it with the code of
VL-Dedup (Section~\ref{sec:benchmark}).
\item We show that labeled pairs hide graph errors, even when they score
clusters. A gradient boosting classifier selects no false edge among the
labeled test pairs, and its co-membership F1 on these pairs is 0.9820. Over
all record pairs, its clusters join 7,711,852 pairs of different entities.
These errors spread along paths of edges, so the size cap repairs them better
than fixes to single edges. The cap lowers the false joins to 385, and over
eight settings it stays within 0.0051 of the best of five clustering
algorithms from ER (Section~\ref{sec:exp-graph}).
\item We show that results on the original records overstate robustness. A
classifier with only image features reaches F1 0.9997 on the original records
and 0 on Strong, the test images with the strongest edits. The cap repairs
the clusters of VL-Dedup on every test surface and on COCO and SOP corpora of
up to 580,767 records. On the pairs of SOP records with several photos per
product (product pairs), VL-Dedup reaches F1 0.2909, against 0.1658 for the
strongest baseline (Section~\ref{sec:exp-graph}).
\item We show that the same VLM judges single records better than pairs, and
that correct clusters still lose facts. For captions taken from other images
(swapped captions), the probability of the VLM for one record reaches F1
0.9264, and the combined record score reaches 0.9429, against 0.6549 for the
pair verdict. Inside clusters, the VLM also finds complementary captions with
higher precision and recall than on pairs with uncertain scores. In 50
sampled test clusters, keeping one record per cluster loses a fact in 70 to
141 of its 187 deletions. If it deleted 28 records at random, it would lose
10.5 to 21.1, against 5 to 15 for the 28 deletions of VL-Dedup. The ranges
span a strict and a lenient count of two AI auditors
(Sections~\ref{sec:exp-actions} and~\ref{sec:exp-retained}).
\end{itemize}

\paragraph{Outline.} Section~\ref{sec:related-work} reviews related work.
Section~\ref{sec:problem} defines graded multimodal ER and its three
evaluation levels. Section~\ref{sec:method} describes VL-Dedup, and
Section~\ref{sec:benchmark} describes VL-DedupBench.
Section~\ref{sec:experiments} reports our experiments.
Section~\ref{sec:discussion} discusses how to apply VL-Dedup and
VL-DedupBench, and Section~\ref{sec:conclusion} concludes.

\section{Related Work}
\label{sec:related-work}

In this section, we review the work that VL-Dedup builds on and compares with:
data deduplication, captions and image--text consistency, entity resolution,
the clustering of pair decisions, and multimodal matching.

\paragraph{Data deduplication.}
Data management research treats the collection and cleaning of training data
as a data quality problem \citep{whang2023datacentric}. Training corpora
contain exact copies, near duplicates, and records that repeat the same
meaning. Hashing and n-gram overlap find repeated surface
forms at scale, and removing them reduces verbatim memorization
\citep{lee2022deduplm}. SemDeDup compares pretrained embeddings to find
records with similar meaning \citep{abbas2023semdedup}. For images,
perceptual hashes find near-identical copies \citep{zauner2010perceptual},
and copy detection embeddings also find copies after crops, rotations, and
color changes \citep{pizzi2022sscd}. For image--text data,
LAION provides a corpus at web scale \citep{schuhmann2022laion5b}, and
Webster et al.\ estimate that about 700 million images of its subset
LAION-2B are duplicates \citep{webster2023laion}. DataComp and MINT-1T study filtering and deduplication choices for multimodal
training data \citep{gadre2023datacomp,awadalla2024mint1t}. Data-Juicer
packages data processing operators, including image deduplication
\citep{chen2024datajuicer}, and SemHash uses compact embeddings for fast
semantic deduplication \citep{vandongen2026semhash}. FairDeDup changes which
records SemDeDup keeps so that the pruned image--text data stays balanced
across sensitive concepts \citep{slyman2024fairdedup}, and EcoDatum combines
several quality operators with a deduplication step to select image--text
training data \citep{xu2025ecodatum}. These methods define redundancy by
similarity and report how many records they remove or how a model trained on
the result performs. They do not decide whether two records show the same
visual entity, and they do not keep a record for the facts that its caption
adds. VL-Dedup uses such hashes and embeddings to retrieve and score
candidate pairs, and our experiments compare it with Data-Juicer, SemHash,
SemDeDup, and caption MinHash.

\paragraph{Captions and image--text consistency.}
Other work adds or rewrites captions instead of deleting records. LaCLIP
rewrites each caption with a language model and samples the original or a
rewritten caption in each training step \citep{fan2023laclip}. VeCLIP and
Recap-DataComp-1B write new captions from the images with vision--language
models \citep{lai2024veclip,li2025recap}, and DreamLIP samples several
sentences of one long caption as positive texts for an image
\citep{zheng2024dreamlip}. These works train better models when an image has
more than one caption or a more detailed caption. Filtering methods score
each record instead. LAION keeps the records whose CLIP similarity passes a
threshold \citep{schuhmann2022laion5b}, Data Filtering Networks learn the
filtering model \citep{fang2024dfn}, T-MARS masks the text inside each
image before it scores the image against its caption \citep{maini2024tmars},
and CAT filters noisy records before pretraining
\citep{radenovic2023filtering}. Other work checks whether a caption correctly
describes its image. SeeTRUE evaluates image--text alignment with human
judgments \citep{yarom2023seetrue}, FineMatch detects and corrects mismatches
in specific aspects of a caption \citep{hua2024finematch}, and LVLM4CEC and
ContextGuard-LVLM use large vision--language models to check whether news
images and text are consistent
\citep{tahmasebi2025lvlm4cec,ma2025contextguard}. These methods judge one
image and its caption at a time. VL-Dedup also scores each record in this way
to flag captions that do not correctly describe their images. Whether a
caption adds facts, however, depends on the other captions of the same visual
entity, so VL-Dedup decides it on the record pairs that share a cluster.

\paragraph{Entity resolution.}
ER decides whether records refer to the same real-world entity. Classic work
treats matching as a decision under uncertainty \citep{fellegi1969record} and
splits the pipeline into blocking, similarity computation, and classification
\citep{elmagarmid2007duplicate,christen2012data,konda2016magellan}. Learned
matchers train neural pair models on labeled pairs
\citep{mudgal2018deepmatcher} or fine-tune pretrained language models on
serialized pairs \citep{li2020ditto}. These matchers are evaluated by F1 on
labeled candidate pairs. Recent work uses large language models
for matching
\citep{narayan2023llmem,peeters2025llmem,fan2024batcher,wang2024comem} and
for clustering \citep{fu2025clustellm}.

\paragraph{Clustering pair decisions.}
Pair decisions become entities only after clustering. Hassanzadeh et al.\
compare clustering algorithms for duplicate detection, such as Center,
Merge-Center, and MCL, and show that the algorithm changes the result
\citep{hassanzadeh2009clustering}. Draisbach et al.\ compare many algorithms
that turn pairwise duplicates into entity clusters and find that no single
algorithm is best in all settings \citep{draisbach2019transforming}.
Menestrina et al.\ compare measures that score a predicted partition
against the true one, including pairwise F1 over all record pairs
\citep{menestrina2010evaluating}. Correlation clustering repairs inconsistent
pair decisions, and KwikCluster gives a simple approximation
\citep{ailon2008aggregating}. Size constraints on
clusters appear in data clustering \citep{zhu2010sizeconstraints} and in
correlation clustering \citep{puleo2015correlation}. VL-Dedup uses such a
bound to stop paths of wrong edges, and our experiments compare it with
Center, Merge-Center, KwikCluster, MCL, and Louvain
\citep{blondel2008louvain}. Other work uses transitivity to infer pair answers
and to save questions to a crowd or an oracle
\citep{wang2013transitive,firmani2016oracle}, and noisy answers need extra
checks to avoid wrong merges \citep{galhotra2018robust}. These answers decide
identity, so each answer can change the clusters. VL-Dedup asks a VLM only
about the record pairs that share a cluster, and its verdicts set caption
actions without changing the clusters.

\paragraph{Multimodal matching.}
Product matchers combine image and text representations to match offers
across shops \citep{wilke2021multimodaler,gvdb2024intermediatefusion}, and an
industry pipeline combines compact multimodal projections with human review
\citep{more2024fashionproduct}. These systems decide product identity. Each
line of work above thus makes only part of the decisions of graded multimodal
ER. Deduplication removes records without deciding visual identity, work on
captions handles each record alone, and ER and multimodal matching decide only
identity. VL-Dedup gives each pair a visual identity decision, a caption
action, and a mismatch action. We know of no benchmark for image--text data
that scores clusters over all record pairs together with the coverage of
deleted records by the retained set. VL-DedupBench scores both, and the
labeled pairs, on the same records.

\section{Problem Definition}
\label{sec:problem}

We first define the records, the actions, and the retained set of graded
multimodal ER, and then the three levels on which we evaluate them.

A corpus $\mathcal{D}=\{r_1,\dots,r_n\}$ contains image--caption records
$r_i=(I_i,c_i)$. Two records can show the same visual entity with captions
that repeat each other, show the same visual entity with captions that add different
facts, or show different entities. A record can also carry a caption that does
not correctly describe its image.

\begin{definition}[Graded multimodal ER]
Given a corpus $\mathcal{D}$, graded multimodal ER assigns each pair
$e=\{r_i,r_j\}$ the action tuple $a_e=(a_e^G,a_e^C,a_e^M)$ defined below,
groups the records into \emph{visual-entity clusters} $\mathcal{C}$, and
writes a retained set $\mathcal{R}\subseteq\mathcal{D}$.
\end{definition}

\begin{itemize}
\item The graph action $a_e^G$ is \texttt{select\_\allowbreak edge} when the
two records show the same visual entity, and \texttt{none} otherwise. The
selected edges form the clusters (Section~\ref{sec:backbone}), and we write
$C(r)$ for the cluster of record $r$.
\item The caption action $a_e^C$ is \texttt{keep\_\allowbreak augment} when
both captions of a pair in one cluster state supported facts that the other
lacks, and \texttt{none} otherwise.
\item The mismatch action $a_e^M$ is \texttt{flag\_\allowbreak review} when a
caption of the pair does not correctly describe its image, and \texttt{none} otherwise.
\end{itemize}

These definitions give the correct actions, and a system predicts them.
A cluster can contain several complementary captions, so cluster membership
alone does not decide which records to delete. An \emph{export policy} turns
the actions of each cluster into record actions (keep, delete, or hold for
review) and writes the retained set $\mathcal{R}$.

We evaluate these decisions on three levels. The benchmark provides labeled
pairs and, for every record, the visual group $g(r)$ of its image
(Section~\ref{sec:benchmark}).

\paragraph{Pair level.}
The pair level scores each action on labeled pairs. Let $\mathcal{T}$ be the
labeled test pairs whose visual relation is known. For a pair
$e=\{r_i,r_j\}\in\mathcal{T}$, the label and the graph decision are
\begin{align}
y_e &= \mathbf{1}\bigl[g(r_i)=g(r_j)\bigr], \label{eq:pair-label}\\
\hat{y}_e &= \mathbf{1}\bigl[a_e^G=\texttt{select\_edge}\bigr]. \label{eq:pair-decision}
\end{align}
A selected edge is the direct decision on its pair. We report precision,
recall, and F1 of $\hat{y}_e$ against $y_e$ over $\mathcal{T}$, as pair
classifiers and deduplication baselines are usually scored. Co-membership
replaces $\hat{y}_e$ with $\mathbf{1}[C(r_i)=C(r_j)]$, so it checks whether
the two records share a cluster. Caption and mismatch actions are scored on the
labeled pairs inside clusters, because an export deletes a record only in
favor of another record of its cluster. Section~\ref{sec:benchmark} gives
their labels.

\paragraph{Graph level.}
The graph level scores the clusters that the selected edges form. Let $\mathcal{D}_P$ be the records of partition $P$, without records with a
\emph{swapped caption}, a caption taken from another image. For all pairs of records in
$\mathcal{D}_P$, we count
\begin{equation}\label{eq:pairwise}
\begin{aligned}
\mathrm{TP} &= \bigl|\{\{r,r'\}\subseteq\mathcal{D}_P: C(r)=C(r'),\ g(r)=g(r')\}\bigr|,\\
\mathrm{FP} &= \bigl|\{\{r,r'\}\subseteq\mathcal{D}_P: C(r)=C(r'),\ g(r)\neq g(r')\}\bigr|,\\
\mathrm{FN} &= \bigl|\{\{r,r'\}\subseteq\mathcal{D}_P: C(r)\neq C(r'),\ g(r)=g(r')\}\bigr|,
\end{aligned}
\end{equation}
and report pairwise precision, recall, and F1, the standard measure for
clusters in ER \citep{menestrina2010evaluating}. We call a pair counted in
$\mathrm{FP}$ a \emph{false join}. The clusters come from the graph over the
whole corpus, and $\mathcal{D}_P$ only selects the pairs we score. Records
with a swapped caption are left out because the right action for them is a
mismatch action, not a graph action.

\paragraph{Retained-set level.}
A deleted record $r\notin\mathcal{R}$ is \emph{covered} when the retained
records of its cluster, alone or together, state every fact of its caption
that its image supports. Otherwise it is \emph{lost}. We report the covered
and lost records of each policy.

\paragraph{Example.}
In Figure~\ref{fig:teaser}(b), the labeled pairs are B1--B2 with $y_e=1$ and
A2--B1 with $y_e=0$. The gradient boosting classifier selects B1--B2 and
rejects A2--B1, so precision and recall on the labeled pairs are both 1. It
also selects A1--A2 and A2--B2, which puts the four records into one cluster.
The graph level leaves out A2, whose caption is swapped. Of the three pairs of
A1, B1, and B2, B1--B2 is a true positive, and A1--B1 and A1--B2 are false
joins, so pairwise
precision is 1/3 and recall is 1. At the retained-set level, an export that
keeps one record per cluster keeps A1, which has the highest CLIPScore of the
four, and deletes A2, B1, and B2. B1 and B2 are lost, because no retained
caption states that a giraffe and a zebra are in a grassy field. A2 is
covered, because its image supports none of the facts in its caption.

\section{VL-Dedup}
\label{sec:method}

\begin{figure*}[!t]
\centering
\includegraphics[width=\textwidth]{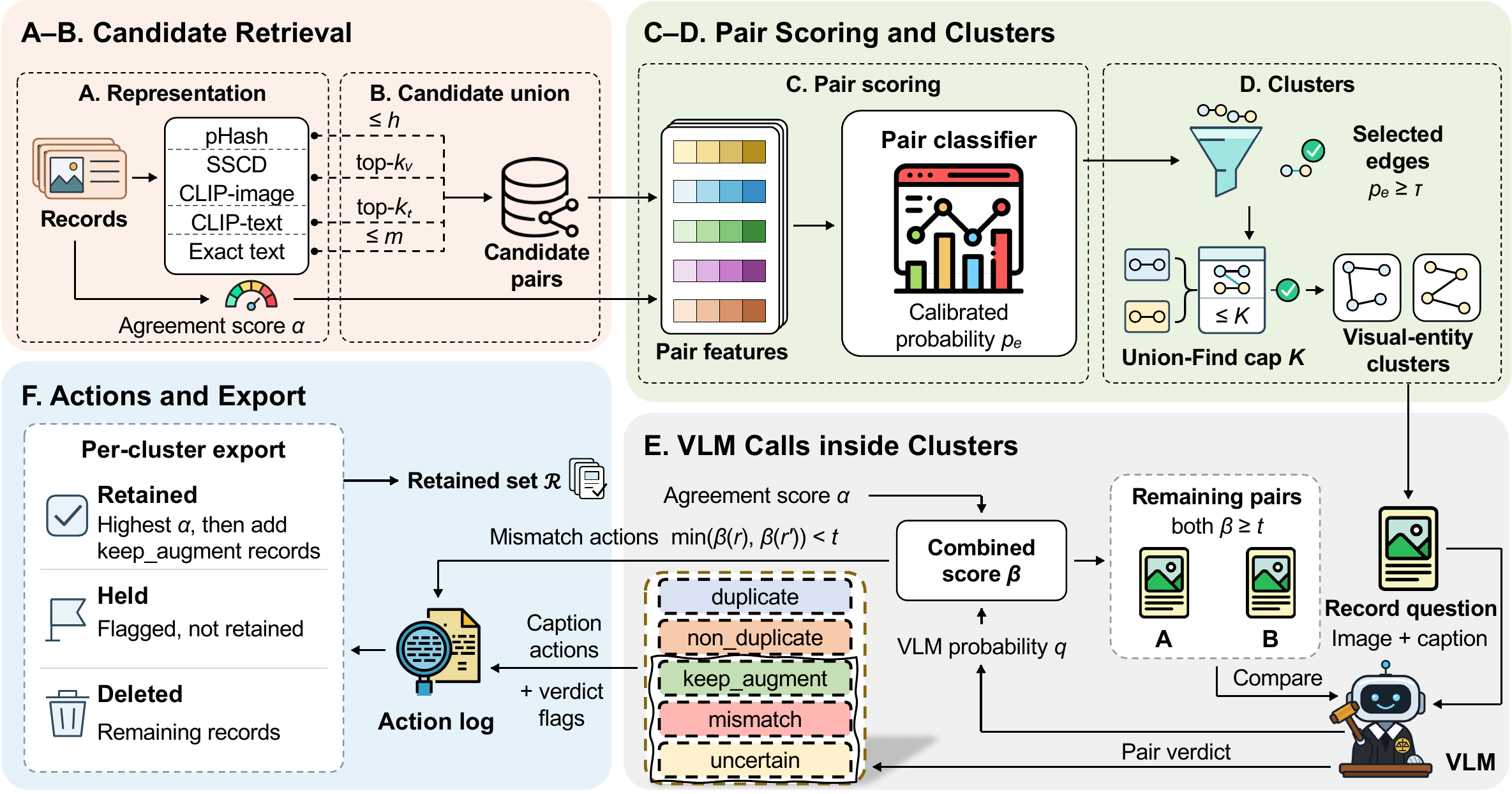}
\caption{Overview of VL-Dedup. Stages A--D build visual-entity clusters of at
most $K$ records. Stages E--F set caption and mismatch actions inside the
clusters and export the retained set.}
\label{fig:pipeline}
\end{figure*}

VL-Dedup separates visual identity from caption and mismatch actions.
Figure~\ref{fig:pipeline} shows its six stages. Stages A--B retrieve candidate
pairs $\mathcal{E}_{\mathrm{cand}}$ (Section~\ref{sec:repr}). Stage~C scores
each pair, and Stage~D selects edges and builds visual-entity clusters under a
size cap (Section~\ref{sec:backbone}). Stage~E asks a VLM about each clustered record and about the remaining pairs
inside each cluster (Section~\ref{sec:vlm}).
Stage~F turns the verdicts and the record scores into caption and mismatch
actions and exports the retained set (Section~\ref{sec:resolution}).
Section~\ref{sec:e2e} analyzes the cost.

\subsection{Candidate Retrieval}\label{sec:repr}

\paragraph{Stage A: representation.}
Stage~A computes four kinds of signals for each record. A perceptual hash of
the image finds near-identical copies by a small Hamming distance. A copy
detection embedding compares images under crops, rotations, and color
changes. Image and caption embeddings in one shared space match records by
content rather than by pixels or words, so an image can also find captions
that describe it. An image--text agreement score $\alpha(r)$ rates how well the
caption of record $r$ matches its own image. A low score marks a possible
mismatch.

\paragraph{Stage B: candidate union.}
For each record, Stage~B queries four channels and takes the union of the
results: images within Hamming distance $h$, the top-$k_v$ image neighbors,
the top-$k_t$ caption neighbors, and up to $m$ records with the same
normalized caption. Each result becomes one undirected candidate pair. Stage~B
stores the channel and score of each pair, and later stages reuse these values
without querying the indexes again. The limit $m$ prevents a common caption
from pairing one record with every record that shares it.

\subsection{Pair Scoring and Clusters}\label{sec:backbone}

\paragraph{Stage C: pair scoring.}
Stage~C describes each candidate pair with 11 features computed from its two
records. They are the image embedding similarity, the caption embedding
similarity, the perceptual hash similarity, the copy detection similarity,
exact equality of the raw and of the normalized captions, the minimum, mean,
and difference of the two agreement scores, and the minimum and mean of the
two cross scores between the image of one record and the caption of the
other. The features use only pixels, captions, and values computed from them.
They use no file name, source identifier, construction grade, or partition.
A pair classifier trained on labeled pairs predicts the relation of each pair. Its
visual score $s_e$ is the total probability of the relations whose two records
show the same visual entity. An isotonic regression fit on validation pairs
maps $s_e$ to a calibrated probability $p_e\in[0,1]$.

\paragraph{Stage D: selected edges and clusters.}
Stage~D selects the edges
\begin{equation}\label{eq:selected}
\mathcal{S}=\{e\in\mathcal{E}_{\mathrm{cand}}: p_e\ge\tau\}.
\end{equation}
The threshold $\tau$ has the highest validation recall among the thresholds
whose validation precision is at least $\pi$.

A wrong edge merges two clusters, and every later edge into either cluster
extends the merge. Paths of such edges join records that no selected edge
compares. Stage~D therefore limits each cluster to at most $K$ records
(Algorithm~\ref{alg:size-cap}). It adds the edges of $\mathcal{S}$ to a
union--find structure from the highest $p_e$ to the lowest and skips an edge
when the merged cluster would exceed $K$ records. Ties in $p_e$ are broken by
the raw score and then by record identifiers.

\begin{algorithm}[t]
\caption{Stage D: clusters under a size cap}
\label{alg:size-cap}
\begin{algorithmic}[1]
\Require records $\mathcal{D}$, selected edges $\mathcal{S}$ with probabilities $p_e$, size cap $K$
\Ensure visual-entity clusters $\mathcal{C}$
\State put each record in its own cluster
\For{$e=\{r,r'\}\in\mathcal{S}$ in decreasing order of $p_e$}
  \If{$C(r)\neq C(r')$ and $|C(r)|+|C(r')|\le K$}
    \State merge $C(r)$ and $C(r')$
  \EndIf
\EndFor
\State \Return the current clusters as $\mathcal{C}$
\end{algorithmic}
\end{algorithm}

We choose $K$ from a grid by the pairwise F1 of Equation~\ref{eq:pairwise} on
the validation partition, and take the smallest $K$ among ties. With
$K=\infty$, $\mathcal{C}$ is the set of connected components of
$(\mathcal{D},\mathcal{S})$. The cap changes only which edges join clusters.
It does not change $p_e$ or $\mathcal{S}$.

\begin{proposition}\label{prop:cap}
For any $K$ and any order of $\mathcal{S}$, every cluster of
Algorithm~\ref{alg:size-cap} has at most $K$ records and lies inside one
connected component of $(\mathcal{D},\mathcal{S})$. Every false join under
the cap is therefore also a false join without the cap, and the number of
pairs inside clusters, that is, the record pairs that share a cluster, is
\begin{equation}\label{eq:vlm-calls}
\sum_{c\in\mathcal{C}}\frac{|c|\,(|c|-1)}{2}\le\frac{n(K-1)}{2}.
\end{equation}
\end{proposition}

\begin{proof}
Algorithm~\ref{alg:size-cap} merges two clusters only along an edge of
$\mathcal{S}$, and only when the merged cluster has at most $K$ records. Each
cluster is therefore connected in $(\mathcal{D},\mathcal{S})$ and lies inside
one component. Two records that share a cluster under the cap thus share a
component without the cap. A cluster $c$ holds
$|c|(|c|-1)/2\le|c|(K-1)/2$ pairs, and the sizes $|c|$ sum to $n$.
\end{proof}

\subsection{VLM Calls inside Clusters}\label{sec:vlm}

A deduplication export deletes a record only in favor of another record of
its cluster. Caption decisions therefore matter for the pairs inside
clusters. We write $U$ for the set of all record pairs that share a cluster,
\begin{equation}\label{eq:review-set}
U=\{\{r,r'\}: r\neq r',\ C(r)=C(r')\}.
\end{equation}

\paragraph{Record scores.}
Stage~E first asks the VLM about each record in a cluster of at least two
records. The VLM receives the image and the caption and answers whether the
caption correctly describes the image. Stage~E reads the probabilities of the
answers yes and no among the 20 most likely first tokens and keeps
$q(r)$, the probability of yes divided by the sum of the two. The combined
score puts $q(r)$ and the agreement score $\alpha(r)$ on one scale,
\begin{equation}\label{eq:combined}
\beta(r)=\tfrac{1}{2}\bigl(F_\alpha(\alpha(r))+F_q(q(r))\bigr),
\end{equation}
where $F_\alpha$ and $F_q$ give the fraction of validation records in
clusters whose score is lower, with ties counted as one half. A low
$\beta(r)$ marks a caption that may not correctly describe its image.

\paragraph{Pair verdicts.}
Pairs whose lower combined score is below the threshold $t$ already receive a
mismatch action (Section~\ref{sec:resolution}), so Stage~E sends only the
other pairs inside clusters to the VLM. The VLM receives both images and both
captions and no label, and returns one pair verdict $z_e$:
\texttt{duplicate}, \texttt{keep\_\allowbreak augment},
\texttt{non\_\allowbreak duplicate}, \texttt{mismatch}, or
\texttt{uncertain}. The first three mean that the images show the same visual
entity and the captions repeat each other, that they show the same visual
entity and the captions add complementary facts, and that the visual entities
differ. A \texttt{mismatch} verdict flags an unsupported caption or
conflicting visible facts, and \texttt{uncertain} covers ambiguous cases and
parse failures. The VLM answers with one structured object whose fields state
whether the images show the same visual entity, whether each caption is
supported by its image, whether a blocker prevents deduplication, and whether
the captions add information to each other. A verdict that contradicts these
fields becomes \texttt{uncertain}.

\subsection{Actions and Export}\label{sec:resolution}

Stage~F assigns each pair its action tuple. The graph action is
\texttt{select\_\allowbreak edge} exactly when $e\in\mathcal{S}$ and its two
records share a cluster. For a pair in $U$, the caption action is
\texttt{keep\_\allowbreak augment} exactly when
$z_e=\texttt{keep\_\allowbreak augment}$. The mismatch action uses the
combined scores,
\begin{equation}\label{eq:mismatch-rule}
a_e^M=\texttt{flag\_\allowbreak review}
\iff \min\bigl(\beta(r),\beta(r')\bigr)<t,
\end{equation}
where $t$ has the highest F1 on the validation pairs inside clusters that carry a
mismatch or same-entity label. Stage~E sends a pair to the VLM only when both
of its records have a combined score of at least $t$. If the VLM still returns
\texttt{mismatch}, a caption of the pair may hold an error that the scores
miss. Stage~F gives such a pair a \emph{verdict flag}, holds it for review, and does
not count it as a mismatch action. The verdicts and flags do not change $\mathcal{S}$
or $\mathcal{C}$.

The export policy keeps one representative per cluster, the record with the
highest agreement score $\alpha(r)$. It then keeps every record that has
a \texttt{keep\_\allowbreak augment} action with a retained record, and repeats
this step until no record is added. A record that is not kept and lies on a
pair with a mismatch action or a verdict flag is held for review. The policy deletes the remaining records.
Stage~F stores the clusters, the pair actions, and the verdict of each
pair sent to the VLM, so each deletion can be traced to the actions behind it.

\subsection{Cost}\label{sec:e2e}

Let $n$ be the number of records and $d$ the total embedding dimension. Exact
nearest-neighbor search costs $O(n^2d)$, and approximate indexes reduce this
cost. The embedding and exact-text channels return a fixed number of
neighbors per record, so their candidates grow linearly with $n$. Scoring is
linear in $|\mathcal{E}_{\mathrm{cand}}|$. Sorting the selected edges costs
$O(|\mathcal{S}|\log|\mathcal{S}|)$, and union--find over them is
near-linear. The size cap also bounds the VLM calls of Stage~E. By
Proposition~\ref{prop:cap}, $|U|\le n(K-1)/2$, and Stage~E adds at most $n$
calls for the record scores. The total cost is
\begin{equation}\label{eq:cost-e2e}
O\!\left(n^2d+|\mathcal{E}_{\mathrm{cand}}|+|\mathcal{S}|\log|\mathcal{S}|+nK\right),
\end{equation}
where the last term counts VLM calls. For a fixed $K$, the VLM calls grow
linearly with $n$. Without the cap, one cluster of $n'$ records alone needs
$n'(n'-1)/2$ calls.

\section{VL-DedupBench}
\label{sec:benchmark}

VL-DedupBench tests visual identity, caption actions, and mismatch actions on
the same records. It contains 50,000 records, 36,000 from Common Objects in
Context (COCO)~\citep{lin2014microsoftcoco} and 14,000 from Stanford Online
Products (SOP)~\citep{song2016deepmetric}, and 66,788 labeled pairs. COCO
records use human captions. SOP provides product photos without captions, so
each SOP caption is generated from its image by InternVL3-2B
\citep{zhu2025internvl3}. Each record has a stable identifier, an image, a caption,
source metadata, a visual group, and a partition. Because every record has a
visual group, the graph level can score all record pairs of a partition, not
only the labeled pairs.
Table~\ref{tab:benchmark-surface-integrity} reports the relation counts. We describe the relations, the partitions, and the
audit of hard negatives, and then the test surfaces
(Section~\ref{sec:surfaces}).

\begin{table}[t]
\centering
\small
\setlength{\tabcolsep}{4pt}
\caption{Relation counts of VL-DedupBench over all partitions (Full) and in
the test partition (Test).}
\label{tab:benchmark-surface-integrity}
\begin{tabular*}{\linewidth}{@{\extracolsep{\fill}}ccc@{}}
\toprule
Relation & Full & Test \\
\midrule
L0/L1/L3 & 52,618 & 7,941 \\
L4 & 5,037 & 760 \\
M & 3,599 & 543 \\
L2 & 4,481 & 634 \\
HN & 1,053 & 159 \\
All & 66,788 & 10,037 \\
\bottomrule
\end{tabular*}
\end{table}

\paragraph{Relations.}
The records of one COCO image form a visual group, and each SOP product forms
its own group. L3 pairs join two records of one COCO image with different
human captions. L0 and L1 records are copies of another record in their
group. Within each COCO group that contains at least three L3 records, we
select one pair deterministically as L4
(\texttt{keep\_\allowbreak augment\_\allowbreak candidate}). Selection uses
record identifiers rather than caption content. L2 pairs reuse an exact
caption across different images. Each M record carries a swapped caption,
which comes from another image. HN pairs join SOP records of the same superclass but different
products. Table~\ref{tab:relation-targets} maps each relation to its
references.

\begin{table}[t]
\centering
\small
\setlength{\tabcolsep}{2pt}
\caption{Constructed pair relations and their action references.}
\label{tab:relation-targets}
\begin{tabular}{@{}p{0.14\linewidth}>{\raggedright\arraybackslash}p{0.35\linewidth}p{0.20\linewidth}p{0.26\linewidth}@{}}
\toprule
Grade & Constructed relation & Same visual entity & Action reference \\
\midrule
L0/L1/L3 & \texttt{duplicate} & yes & none \\
L4 & \texttt{keep\_\allowbreak augment\_\allowbreak candidate} & yes & caption labels \\
L2 & \texttt{text\_\allowbreak duplicate\_\allowbreak only} & no & none \\
HN & \texttt{hard\_\allowbreak negative} & no & none \\
M & \texttt{mismatch} & not scored & \texttt{flag\_\allowbreak review} \\
\bottomrule
\end{tabular}
\end{table}

\paragraph{Partitions.}
COCO records share a split key when they use the same source image, and SOP
records use their superclass identifier. A deterministic hash of the split key
assigns it to training, validation, or test with cut points 0.70 and 0.85.
The benchmark has 7,208 split keys, and no visual group, source image, or
labeled pair crosses partitions. The record split is 35,820/7,639/6,541, and
the pair split is 46,425/10,326/10,037.

\paragraph{Audit of hard negatives.}
Two AI auditors, GPT-5.6-terra and Claude Opus 4.8, judged all 1,053 HN pairs
from the visible images and captions with a fixed five-label guide. The inputs
hid classes, grades, predictions, and the other auditor's answer. The 32
pairs with a disagreement or an \texttt{uncertain} answer went to a human,
who kept all 1,053 pairs as \texttt{non\_\allowbreak duplicate}.

\subsection{Test Surfaces}
\label{sec:surfaces}

\begin{figure}[t]
\centering
\includegraphics[width=\columnwidth]{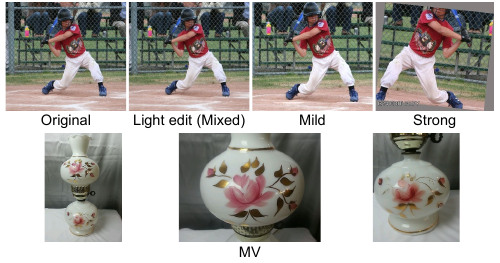}
\caption{Edited copies of one test image and three photos of one test
product in MV.}
\label{fig:surfaces}
\end{figure}

The records of one COCO group share an image file. To test identity decisions
when the image bytes differ, we add three kinds of test surfaces. All
models, calibrators, thresholds, and $K$ stay fixed on every surface.
Table~\ref{tab:surfaces} gives the test populations, and
Figure~\ref{fig:surfaces} shows examples. The validation partition has 7,079
records, 25,052,581 record pairs, and 8,959 same-entity pairs.

\paragraph{Edited copies.}
Each edit writes a new image file for the 5,709 test records that occur in
labeled pairs. Mild applies a 12\% center crop, JPEG quality 70, and a
brightness change of up to 8\%. Strong applies a 35\% crop, a 6$^\circ$ rotation, a horizontal flip on half of the records, a
color shift, and a text watermark. Mixed replaces a random 2,901 of the 5,709
images with a light edit (JPEG quality 85 and 0.9 scaling), so a same-entity
pair can join an original and an edited image. The 5,709 records
come from 1,362 distinct images, and two AI auditors check one edited copy of
each image in Mild and in Strong. Neither finds the subject removed in any
Mild copy. In Strong, both find it removed in 5 copies, and at least one in
10.

\paragraph{Several photos of one product (MV).}
SOP has several photos of each product, and VL-DedupBench keeps one. MV
adds two more photos for each validation and test product and for 1,500
training products, 8,550 records in total. Each added record receives a
caption generated in the same way and the visual group of its product. MV adds
2,984 same-product test pairs and 2,984 sampled pairs of different products
to the labeled pairs. The photos of one product differ in viewpoint and
background, so copy detection alone does not match them.

\paragraph{Larger entities.}
COCO groups have at most seven records. We extend 300 groups in each of the
validation and test partitions to 10, 20, and 40 records, 100 groups for each
size, with edited copies of the group image (12\% crop, JPEG quality 70, and a
small brightness change) and the existing captions of the group. This adds
5,500 records and 98,500 same-entity pairs to each partition.

\paragraph{Action references.}
The L4 label marks a pair as a candidate for
\texttt{keep\_\allowbreak augment}. To score caption actions, we label
caption content directly on 400 test pairs inside clusters, 200 from L3 and 200 from L4. A
pair is complementary when the image supports both captions and each caption
states a fact that the other lacks. Two AI auditors label every pair, and a
human decides the pairs on which they disagree. Mismatch actions are scored on
the labeled pairs inside clusters, with M pairs as positives and same-entity
pairs as negatives. Retained sets are scored by checking each deleted record
against the full retained set of its cluster (Section~\ref{sec:exp-retained}).

\section{Experiments}
\label{sec:experiments}
\begin{table}[t]
\centering
\small
\setlength{\tabcolsep}{3pt}
\caption{Test populations of the graph level. Records exclude swapped captions.}
\label{tab:surfaces}
\begin{tabular}{@{}lrrrr@{}}
\toprule
Test surface & Records & Record pairs & Same-entity & Labeled \\
\midrule
Original & 5,998 & 17,985,003 & 8,701 & 9,494 \\
Mixed, Mild, Strong & 5,998 & 17,985,003 & 8,701 & 9,494 \\
MV & 8,044 & 32,348,946 & 11,685 & 15,462 \\
Larger entities & 11,498 & 66,096,253 & 107,201 & 9,494 \\
\bottomrule
\end{tabular}
\end{table}

We present the results of VL-Dedup on VL-DedupBench and on two full corpora
outside it. Section~\ref{sec:exp-graph} scores visual identity on the labeled
pairs and over all record pairs. Section~\ref{sec:exp-actions} scores caption
and mismatch actions on the labeled pairs inside clusters.
Section~\ref{sec:exp-retained} scores the retained set, and
Section~\ref{sec:exp-cost} measures cost and scale. We found the following.
(1) Labeled pairs hide graph errors, even when they score clusters. A
gradient boosting classifier selects no false edge among the labeled test
pairs, and its co-membership F1 on these pairs is 0.9820. Over all record
pairs, its clusters join 7,711,852 pairs of different entities, and pairwise
F1 is 0.0023.
(2) Graph errors spread along paths of edges, so a limit on cluster size
repairs them better than fixes to single edges. Removing one edge separates
only 147 of the 307 false joins of VL-Dedup without the cap. Over eight
settings, the cap stays within 0.0051 of the best of five clustering
algorithms from ER, while MCL falls up to 0.18 below the best and KwikCluster
up to 0.33.
(3) Results on the original records overstate robustness. A classifier with
only image features reaches F1 0.9997 on the original records and 0 on
Strong. A stricter edge threshold raises F1 on the original records to 0.9984
but lowers it to 0.1774 on product pairs.
(4) The same VLM judges single records better than pairs, and pairs inside
clusters better than pairs with uncertain scores. For swapped captions, its
probability for one record reaches F1 0.9264, against 0.6549 for its pair
verdict, and the combined score with CLIPScore reaches 0.9429. Inside
clusters, its caption actions gain 13.1 points of precision and 13.6 points
of recall. A pair with a verdict flag still holds a wrong caption 1.7 to 4.1
times as often as a pair that the VLM does not flag.
(5) Correct clusters still lose facts, and VL-Dedup loses fewer because it
chooses which records to delete, not only because it deletes fewer. Keeping
one record per cluster loses a fact in 70 to 141 of its 187 deletions. If it
deleted 28 records at random, it would lose 10.5 to 21.1 in expectation,
against 5 to 15 for the 28 deletions of VL-Dedup.

\subsection{Setup}
\label{sec:exp-setup}

\paragraph{Implementation.}
Stage~A uses pHash \citep{zauner2010perceptual}, SSCD copy detection
embeddings \citep{pizzi2022sscd}, CLIP image and text embeddings
\citep{radford2021clip}, and CLIPScore as the agreement score
\citep{hessel2021clipscore}. Stage~B uses $h=2$, $k_v=30$, $k_t=10$, and
$m=20$, which gives 1,502,683 candidates over the 50,000 records and 1,771,452
over the 58,550 records with MV. Stage~C fits logistic regression to the 11
features on the 46,425 training pairs. With $\pi=0.99$, the calibrated
threshold is $\tau=0.6667$. Validation selects $K=5$ for Stage~D and
$t=0.0596$ for the combined score. Stage~E uses Qwen2.5-VL-7B
\citep{bai2025qwen25vl} at temperature 0, with one output token for the
record question and at most 384 output tokens for the pair verdict. All
settings come from training and validation data.

\paragraph{Baselines.}
B2 thresholds the SSCD cosine similarity \citep{pizzi2022sscd}. B6 fits
balanced logistic regression to SSCD cosine, SimCSE cosine
\citep{gao2021simcse}, and exact equality of normalized captions. Both choose
their thresholds by pair F1 on validation, and their clusters are the
connected components of the selected pairs. Data-Juicer v1.5.3 runs its pHash
image deduplicator \citep{chen2024datajuicer}. SemHash v0.4.1 runs
self-deduplication on CLIP image and text embeddings, with threshold 0.9
chosen on validation pairs \citep{vandongen2026semhash}. SemDeDup
\citep{abbas2023semdedup} groups the CLIP image embeddings with $k$-means and
marks two records of one group as duplicates when their cosine similarity is
at least $1-\varepsilon$. Validation selects 50 $k$-means clusters and $\varepsilon=0.005$.
Data-Juicer also runs its MinHash deduplicator on the captions with its
default settings. We evaluate the duplicate groups that each system outputs. We also replace the classifier of
Stage~C with gradient boosting (HGB) on the same 11 features.

\subsection{From Pairs to Clusters}
\label{sec:exp-graph}

The graph stage of VL-Dedup has two parts, the pair classifier of Stage~C and the
size cap of Stage~D. We first show that labeled pairs hide the errors that
clustering adds, and that the cap repairs them for either classifier. We then
compare the clusters with deduplication tools and copy detectors on every
test surface. Table~\ref{tab:graph} gives pairwise F1 on the labeled test
pairs and over all test record pairs and, for each relation, the labeled test pairs whose two records share
a cluster.

\begin{table*}[t]
\centering
\footnotesize
\setlength{\tabcolsep}{2pt}
\caption{Visual identity on the test surfaces. The left block gives pairwise F1
(Equation~\ref{eq:pairwise}) over the 9,494 labeled test pairs (Labeled) and
over all test record pairs. The right block counts the labeled test pairs of
each relation whose two records share a cluster, on the original records and,
for the product columns, on MV. $\uparrow$ marks pairs that should share a
cluster, and $\downarrow$ pairs that should not. Bold marks the best F1 in
each column.}
\label{tab:graph}
\begin{tabular*}{\textwidth}{@{\extracolsep{\fill}}llcccccccccccccc@{}}
\toprule
\multirow{2}{*}{Method} & \multirow{2}{*}{Modality} & \multirow{2}{*}{$K$} & \multicolumn{7}{c}{Pairwise F1} & \multicolumn{6}{c@{}}{Labeled test pairs in one cluster} \\
\cmidrule(lr){4-10}\cmidrule(l){11-16}
 & & & Labeled & Original & Mixed & Mild & Strong & MV & Products & L0/L1/L3\,$\uparrow$ & L4\,$\uparrow$ & Same product\,$\uparrow$ & L2\,$\downarrow$ & HN\,$\downarrow$ & Two products\,$\downarrow$ \\
\midrule
Data-Juicer & Image & -- & \textbf{1.0000} & \textbf{1.0000} & 0.9800 & 0.9054 & 0.5997 & 0.8572 & 0.0420 & 7,941 & 760 & 64 & 0 & 0 & 0 \\
SemDeDup & Image & -- & \textbf{1.0000} & \textbf{1.0000} & 0.6672 & 0.6910 & 0.1568 & 0.8555 & 0.0219 & 7,941 & 760 & 33 & 0 & 0 & 0 \\
B2 SSCD & Image & $\infty$ & \textbf{1.0000} & 0.9998 & \textbf{0.9998} & \textbf{0.9998} & 0.9979 & 0.8611 & 0.0900 & 7,941 & 760 & 141 & 0 & 0 & 0 \\
\addlinespace[2pt]
Caption MinHash & Text & -- & 0.0679 & 0.0722 & 0.0722 & 0.0722 & 0.0722 & 0.0614 & 0.0290 & 327 & 1 & 44 & 634 & 0 & 0 \\
\addlinespace[2pt]
SemHash & Image+text & -- & 0.5164 & 0.3753 & 0.3494 & 0.4100 & 0.3599 & 0.3360 & 0.1495 & 2,791 & 243 & 259 & 15 & 0 & 0 \\
B6 fusion & Image+text & $\infty$ & \textbf{1.0000} & 0.9994 & 0.9994 & 0.9994 & \textbf{0.9995} & 0.8668 & 0.1658 & 7,941 & 760 & 274 & 0 & 0 & 0 \\
\midrule
HGB & Image+text & $\infty$ & 0.9820 & 0.0023 & 0.0023 & 0.0022 & 0.0022 & 0.0017 & 0.0039 & 7,939 & 759 & 576 & 310 & 6 & 157 \\
HGB & Image+text & 5 & 0.9995 & 0.9779 & 0.9772 & 0.9779 & 0.7256 & 0.8242 & 0.1495 & 7,934 & 759 & 338 & 0 & 0 & 1 \\
VL-Dedup & Image+text & $\infty$ & \textbf{1.0000} & 0.9827 & 0.9818 & 0.9859 & 0.9771 & 0.7386 & 0.1532 & 7,941 & 760 & 595 & 0 & 0 & 1 \\
VL-Dedup & Image+text & 5 & 0.9997 & 0.9937 & 0.9937 & 0.9944 & 0.9947 & \textbf{0.8691} & \textbf{0.2909} & 7,936 & 760 & 572 & 0 & 0 & 0 \\
\bottomrule
\end{tabular*}
\end{table*}

\paragraph{Labeled pairs hide graph errors.}
HGB selects no false edge among the 9,494 labeled test pairs and
misses three. Its clusters put 316 labeled pairs of different entities
together, and co-membership F1 on the labeled pairs is 0.9820. Over all
17,985,003 test record pairs, its clusters join 7,711,852 pairs of different
entities, and pairwise F1 is 0.0023 (Table~\ref{tab:graph}). One cluster holds
33,362 records. The classifier of VL-Dedup decides every labeled pair
correctly, both as a selected edge and as co-membership, yet its clusters join
307 pairs of different entities, in clusters of up to 598 records. Almost none
of the falsely joined pairs is labeled, so a benchmark of labeled pairs
cannot see these errors.

\begin{figure}[t]
\centering
\includegraphics[width=\columnwidth]{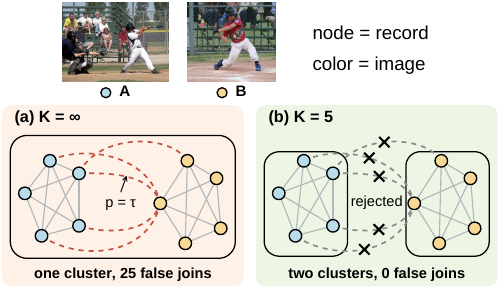}
\caption{One component of the VL-Dedup graph (a) without the cap and (b)
with $K=5$.}
\label{fig:graph-example}
\end{figure}

\paragraph{The size cap.}
Validation selects $K=5$ for both classifiers. On the original records,
the cap lowers the false joins of HGB from 7,711,852 to 385 and raises its F1
from 0.0023 to 0.9779. For VL-Dedup, the false joins fall from 307 to 106 and
F1 rises from 0.9827 to 0.9937. The cap splits the few test entities with
more than five records, which costs 5 missed pairs. All 106 remaining false
joins of VL-Dedup join photos of two different lamp products, the SOP
superclass of the test partition, in 40 clusters, and none joins two COCO
images. The photos are not copies: their pHash values differ in 20 of 64 bits
at the median, and only 4 of the 106 pairs pass the SSCD threshold of B2. On
the full SOP corpus of Section~\ref{sec:exp-cost}, 97\% of the false joins
likewise join two products of one superclass.
Figure~\ref{fig:graph-example} shows one component that the cap splits. Two
images with five records each are linked by five selected edges whose
calibrated probability equals $\tau$. Without the cap, the ten records form
one cluster with 25 false joins. With $K=5$, the edges inside each image
join first, and the cap rejects all five edges between the images.

\begin{figure}[t]
\centering
\includegraphics[width=\columnwidth]{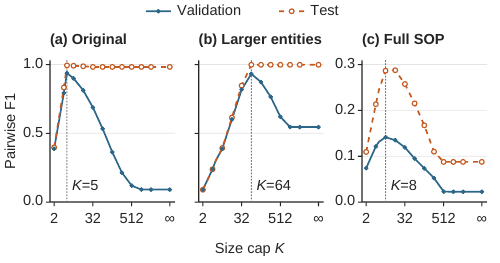}
\caption{Pairwise F1 as the size cap $K$ changes. Dotted lines mark the
values of $K$ chosen on validation.}
\label{fig:k-sensitivity}
\end{figure}

Figure~\ref{fig:k-sensitivity} shows F1 as $K$ changes. On the
original records, F1 is highest at $K=5$ on both validation and test, and it
falls as larger caps let paths of wrong edges grow. Without the cap, F1 is
0.0910 on validation and 0.9827 on test. On the full SOP corpus, validation
selects $K=8$, and test F1 at $K=8$ is 0.0010 below its highest value, at
$K=16$. Every precision target $\pi$ up to 0.9997 selects the same
$\tau=2/3$, because this calibrated probability already accepts every
retrieved same-entity validation pair at precision 0.9998. Only $\pi=1$ gives
$\tau=1$, which raises F1 on the original records to 0.9984 but lowers it to
0.9304 on Strong and to 0.1774 on product pairs.

\paragraph{Copies and edited copies.}
B2 and B6 are copy detectors whose thresholds we choose on validation.
Copies and edited copies are the case they are built for, and they stay
above 0.997 on the original records and on Mixed, Mild, and Strong. The classifier of
VL-Dedup stays within 0.0061 of them. With the cap, VL-Dedup keeps F1 between
0.9937 and 0.9947 on the edited surfaces, against 0.9771 to 0.9859 without
the cap. HGB with the cap falls to 0.7256 on Strong, because the edits push many
same-entity pairs below its threshold. Data-Juicer falls from 1.0000 on the
original records to 0.9800, 0.9054, and 0.5997 as the edits get stronger,
because pHash misses 341, 1,504, and 4,975 same-entity pairs. SemDeDup falls
from 1.0000 to 0.6672, 0.6910, and 0.1568, because the threshold chosen on
validation joins only images whose embeddings are almost identical. SemHash
stays below 0.41 on every surface. Caption MinHash reaches 0.0722 on every
surface, because it finds only the 328 same-entity test pairs whose captions
are nearly identical, and it joins all 634 L2 pairs, whose identical captions
describe different images.

\paragraph{Several photos of one product.}
MV is harder for every method, because two photos of one product can differ
in viewpoint and background. Over all test records with MV, VL-Dedup with the cap
reaches F1 0.8691, which is 0.0022 above B6, 0.0080 above B2, 0.0119 above
Data-Juicer, and 0.0136 above SemDeDup. The difference lies in the product
pairs, the pairs of SOP records in MV, shown in the Products column of Table~\ref{tab:graph}. There, VL-Dedup
reaches F1 0.2909 and finds 572 same-product pairs, against 0.1658 and 274 for B6 and 0.0900 and 141 for B2.
On the labeled product pairs, it joins none of the 2,984 pairs of two
products.
At the thresholds chosen on validation, B2 and B6 join few photos of one
product. Data-Juicer finds 64 such pairs and SemDeDup 33. Without the cap, it joins 4,190 pairs of different products, in clusters of
up to 1,118 records. The cap lowers this to 376 and raises F1 on the product
pairs from 0.1532 to 0.2909. Training with the MV
photos raises F1 on the product pairs to 0.4690 but misses 191 labeled
same-entity pairs on the original records, against 5, so we train without them.

\paragraph{Larger entities.}
When 300 test groups grow to 10, 20, and 40 records, validation on the
extended groups selects $K=64$. VL-Dedup then reaches F1 0.9981 on the
extended test records, with 407 false joins and no missed pair. Without the
cap, F1 stays the same, but the largest cluster grows from 64 to 598 records
and the pairs inside clusters grow from 312,229 to 654,615. With $K=5$, the
cap splits the larger entities and F1 falls to 0.2994. The cap must therefore
be chosen on validation data whose entities have the sizes of the corpus.

\begin{table}[t]
\centering
\footnotesize
\setlength{\tabcolsep}{1.4pt}
\caption{Pairwise test F1 of clustering algorithms on the same selected edges.
Larger denotes the larger entities. Bold marks the best value in each
column.}
\label{tab:clustering}
\begin{tabular}{@{}lcccccccc@{}}
\toprule
& \multicolumn{4}{c}{Edges of VL-Dedup} & \multicolumn{4}{c}{Edges of HGB} \\
\cmidrule(lr){2-5}\cmidrule(l){6-9}
Algorithm & Orig. & Strong & MV & Larger & Orig. & Strong & MV & Larger \\
\midrule
No cap (components) & 0.9827 & 0.9771 & 0.7386 & 0.9981 & 0.0023 & 0.0022 & 0.0017 & 0.0061 \\
Center & 0.5818 & 0.5382 & 0.5116 & 0.1328 & 0.5989 & 0.4862 & 0.4984 & 0.1388 \\
Merge-Center & 0.9827 & 0.9782 & 0.7446 & 0.9986 & 0.0024 & 0.0024 & 0.0018 & 0.0063 \\
KwikCluster & 0.9914 & 0.9817 & 0.8672 & 0.9657 & 0.7241 & 0.4708 & 0.5877 & 0.5563 \\
Louvain & 0.9827 & 0.9771 & 0.7920 & 0.9981 & 0.3045 & 0.2361 & 0.1996 & 0.7450 \\
MCL & \textbf{0.9953} & 0.9879 & \textbf{0.8741} & \textbf{0.9996} & 0.8063 & 0.5463 & 0.6409 & 0.8689 \\
Size cap & 0.9937 & \textbf{0.9947} & 0.8691 & 0.9981 & \textbf{0.9779} & \textbf{0.7256} & \textbf{0.8242} & \textbf{0.8835} \\
\bottomrule
\end{tabular}
\end{table}

\paragraph{Other clustering algorithms.}
Table~\ref{tab:clustering} compares the size cap with clustering algorithms
from ER on the same selected edges. Center and Merge-Center
\citep{hassanzadeh2009clustering}, KwikCluster \citep{ailon2008aggregating},
MCL \citep{vandongen2000mcl}, and Louvain \citep{blondel2008louvain} take
their parameters from the validation data of each surface, as the cap does.
On the edges of VL-Dedup, MCL is 0.0016 above the cap on the original
records, 0.0051 above on MV, and 0.0015 above on the larger entities, and it
is 0.0069 below on Strong. On the HGB edges, the cap is best on all four
surfaces, and MCL is 0.17 to 0.18 below it on the original records, Strong, and
MV. Over the eight settings, the cap is never more than 0.0051 below the best
algorithm. MCL falls up to 0.18 below the best, KwikCluster up to 0.33, and
Louvain up to 0.67. Only the cap bounds the pairs inside clusters by
Equation~\ref{eq:vlm-calls}. On the original records, the largest cluster
has 22 records under MCL, 128 under Louvain, and 585 under Merge-Center,
against 5 under the cap. On the original graph, it takes 0.07 seconds,
against 9.8 seconds for MCL and 4.7 seconds for Louvain. Removing one edge
can separate 147 of the 307 false joins of VL-Dedup without the cap, and
14\% of those of HGB. The other false pairs are joined by several paths, and
the size cap limits all of these paths at once.

\begin{table}[t]
\centering
\small
\setlength{\tabcolsep}{3pt}
\caption{Pairwise test F1 when the classifier of Stage~C is retrained without one
group of features. Each row selects $\tau$ and $K$ on validation.}
\label{tab:ablation}
\begin{tabular}{@{}lccccc@{}}
\toprule
Features & $\tau$ & $K$ & Original & Strong & MV \\
\midrule
All 11 & 0.667 & 5 & 0.9937 & 0.9947 & 0.8691 \\
No copy detection & 0.250 & 5 & 0.9551 & 0.8965 & 0.8102 \\
No perceptual hash & 0.286 & 5 & 0.9605 & 0.9627 & 0.8593 \\
No image embedding & 0.500 & 5 & 0.9908 & 0.9928 & 0.8634 \\
No caption & 0.727 & 5 & 0.9969 & 0.9953 & 0.8703 \\
No cross-modal & 0.250 & 5 & 0.9953 & 0.9951 & 0.8697 \\
Image only & 1.000 & 8 & 0.9997 & 0.0000 & 0.8539 \\
Caption only & 0.467 & 5 & 0.2867 & 0.2818 & 0.2673 \\
\bottomrule
\end{tabular}
\end{table}

\paragraph{Features and retrieval channels.}
Table~\ref{tab:ablation} removes one group of the 11 features at a time and
retrains the classifier with the same pairs, calibration, and selection
rules. The caption group holds the caption embedding similarity and the two
caption equality features. The cross-modal group holds the three CLIPScore
features and the two cross scores. The copy detection similarity matters most. Without it, F1 falls to
0.9551 on the original records and to 0.8965 on Strong. Without the perceptual
hash, F1 falls to 0.9605 and 0.9627. The caption features and the
cross-modal features can replace each other. Removing either group changes
F1 by less than 0.004 on every surface. Removing both leaves only image
features, which give F1 0.9997 on the original records but 0 on Strong, where
the classifier joins no test pair. Caption features alone reach at most
0.29. In Stage~B, the image embedding channel alone retrieves all 8,701
same-entity test pairs on the original records and on Strong. On MV, it
retrieves 10,996 of 11,685, and the other three channels add 34.

\subsection{Caption and Mismatch Actions}
\label{sec:exp-actions}

The actions apply to the pairs inside the clusters of VL-Dedup, 10,290 of
them in the test partition. Table~\ref{tab:examples} shows test records
inside clusters and the actions that VL-Dedup takes on them.

\begin{table}[t]
\centering
\footnotesize
\setlength{\tabcolsep}{3pt}
\caption{Test records inside clusters of VL-Dedup and the actions it takes on them.}
\label{tab:examples}
\begin{tabular}{@{}>{\raggedright\arraybackslash}p{0.68\columnwidth}>{\raggedright\arraybackslash}p{0.29\columnwidth}@{}}
\toprule
Reference and captions & Action of VL-Dedup \\
\midrule
\textit{Complementary.} ``Birds are flying past a steeple with a clock on
it.'' / ``A green and white church tower under a partly cloudy sky.'' &
\texttt{keep\_\allowbreak augment}: keep both \\
\addlinespace
\textit{Redundant.} ``a large plane is sitting in the dirt'' / ``A large
plane is parked in a dirt field.'' & \texttt{duplicate}: keep one \\
\addlinespace
\textit{Swapped caption} on a photo of a bride and groom: ``A man is riding a
skateboard down a road.'' & $\beta=0.020<t$: flag for review \\
\addlinespace
\textit{Wrong object}, confirmed by a human: ``a couple zebra drinking water
out in the desert'' / ``Some giraffes and some big elephants by the water.''
& \texttt{mismatch} with $\beta\geq t$: review \\
\addlinespace
\textit{Lost under keep-one.} Deleted: ``a long hair cat laying on a white
bed''. Kept: ``A cat sleeping on a bed with a small TV in a bedroom.'' &
\texttt{keep\_\allowbreak augment}: keep both \\
\bottomrule
\end{tabular}
\end{table}

\paragraph{Caption actions inside clusters.}
On the 400 labeled pairs inside clusters, 269 are complementary. The VLM
assigns \texttt{keep\_\allowbreak augment} to 217 pairs, and 163 of them are
complementary, so its precision is 0.7512 and its recall is 163/269 = 0.6059.
Precision and recall are 0.7207 and 0.6015 on the L3 pairs, and 0.7830 and
0.6103 on the L4 pairs. We repeat the study on 400 pairs that a routing rule
sends to the VLM because their calibrated score is uncertain, with the same
model, prompt, and reference rule. There, the VLM assigns
\texttt{keep\_\allowbreak augment} to 50 pairs, and 31 of them are
complementary, for precision 31/50 = 0.6200 and recall 31/66 = 0.4697. Inside clusters,
precision is 13.1 points higher and recall is 13.6 points higher.

\begin{table}[t]
\centering
\small
\setlength{\tabcolsep}{3pt}
\caption{Mismatch detection on the labeled test pairs inside clusters, with 372 M
pairs as positives and 8,696 same-entity pairs as negatives. Bold marks the
best precision, recall, and F1.}
\label{tab:mismatch}
\begin{tabular}{@{}lrrrrrr@{}}
\toprule
Rule & TP & FP & FN & P & R & F1 \\
\midrule
CLIPScore $\alpha$ & 324 & 38 & 48 & 0.8950 & 0.8710 & 0.8828 \\
VLM probability $q$ & 340 & 22 & 32 & 0.9392 & 0.9140 & 0.9264 \\
Combined $\beta$ & 347 & 17 & 25 & \textbf{0.9533} & 0.9328 & \textbf{0.9429} \\
\midrule
Pair verdict & 334 & 314 & 38 & 0.5154 & 0.8978 & 0.6549 \\
$\beta$, then verdict & 361 & 323 & 11 & 0.5278 & \textbf{0.9704} & 0.6837 \\
\bottomrule
\end{tabular}
\end{table}

\paragraph{Swapped captions.}
Table~\ref{tab:mismatch} compares mismatch rules on the labeled test pairs
inside clusters. A score rule flags a pair when the lower score of its two
records is below a threshold chosen on validation. The pair verdict finds 334 of the 372 M pairs (swapped captions)
but also flags 314 same-entity pairs, so its F1 is 0.6549. Asked about one record at a time,
the same VLM gives a probability $q$ that reaches F1 0.9264. CLIPScore
reaches 0.8828. The combined score of VL-Dedup reaches 0.9429. It misses 25 M
pairs and flags 17 same-entity pairs, against 48 and 38 for CLIPScore. The
gain comes from the form of the decision. A score for each record, with a
threshold chosen on validation, separates swapped captions far better than a
verdict on each pair. Scores that use the other records of the cluster do not
help. CLIPScore minus the median of its cluster reaches 0.8641, and the mean
SimCSE similarity of a caption to the other captions of its cluster reaches
0.7589. Adding the verdict flags to the combined score finds 14 more M pairs and adds 306 flags on same-entity pairs. The combined
score also removes 1,405 of the 10,290 test pairs from the pair verdicts.

\paragraph{Errors in natural captions.}
The VLM flags 306 same-entity test pairs above $t$ as mismatched. To
see how many of them hold a real error, we audit 59 of these pairs,
sampled by grade, and 60 control pairs above $t$ without such a verdict,
mixed and blind to the group. Two AI auditors, GPT-6.1-sol and
GPT-5.6-terra, label each caption as correct, locally wrong, or wholly wrong. The strict
count calls a caption wrong when both auditors flag it, and the lenient count
when either does. Under the strict count, 34\% of the flagged pairs and 8\%
of the control pairs hold a wrong caption. Under the lenient count, the
shares are 46\% and 27\%. A human check of the pairs on which the auditors
disagree gives 34\% and 12\%, inside this range. Of the 22 captions that both
auditors flag in the flagged group, 17 name a wrong object and 10 a wrong
location.

The combined score therefore sets the mismatch action, because it is precise
on swapped captions. Verdict flags go to review, because
a flagged pair holds a wrong caption 1.7 to 4.1 times as often as a control
pair.

\subsection{Retained Set}
\label{sec:exp-retained}

We sample 50 test clusters, 10 of 2--4 records and 40 of 5 records, 237
records in total, and compare two export policies. The first policy, \emph{keeping one record per cluster}
(keep-one), keeps the record of each cluster with the highest CLIPScore
$\alpha$ and deletes the other 187. The export of VL-Dedup keeps 183 records, holds 26 for review, and
deletes 28. For every deleted record, the two AI auditors compare it with
the full retained set of its cluster. They see no policy, relation, or score.
The strict count calls a deleted record lost when both auditors find a fact
of its caption that no retained record states, and the lenient count when
either does.

\begin{figure}[t]
\centering
\includegraphics[width=\columnwidth]{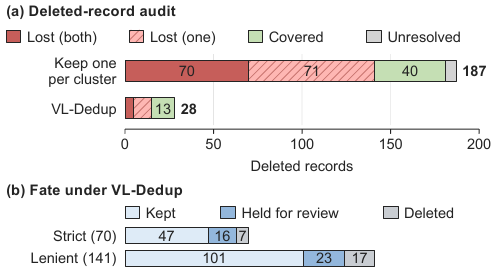}
\caption{Retained-set audit of 50 test clusters. (a) Deleted records by
auditor outcome. (b) What the VL-Dedup export does with the records that
keep-one loses.}
\label{fig:retained}
\end{figure}

Keep-one loses a fact in 70 to 141 of its 187
deletions (Figure~\ref{fig:retained}). The VL-Dedup export loses a fact in 5
to 15 of its 28 deletions. In the clusters of five records, keep-one loses a
fact in 59 to 121 of its 160 deletions, and VL-Dedup in 4 to 11 of its 21.
It deletes fewer records because it keeps records
with complementary captions and holds flagged records for review. If keep-one
deleted only 28 records, chosen at random among its 187 deletions, it would
lose a fact in 10.5 to 21.1 of them in expectation, against 5 to 15 for
VL-Dedup. Of the 70
to 141 records that keep-one loses, VL-Dedup keeps 47 to 101 and
holds 16 to 23 for review.

\subsection{Cost and Scale}
\label{sec:exp-cost}

\begin{table}[t]
\centering
\small
\setlength{\tabcolsep}{2.5pt}
\caption{Measured cost of Stages B--D (median of three runs) and the pairs inside
clusters, $|U|$, with $K=5$ and without a cap.}
\label{tab:cost}
\begin{tabular}{@{}rrrrrrr@{}}
\toprule
Records & Candidates & B (s) & C (s) & D (s) & $|U|$, cap & $|U|$, no cap \\
\midrule
10K & 289,200 & 6.2 & 3.4 & 0.05 & 3,063 & 7,127 \\
20K & 595,089 & 15.4 & 12.9 & 0.11 & 11,970 & 33,901 \\
30K & 900,756 & 28.6 & 18.1 & 0.18 & 26,356 & 100,387 \\
50K & 1,502,683 & 66.7 & 18.8 & 0.41 & 71,930 & 457,215 \\
\bottomrule
\end{tabular}
\end{table}

Table~\ref{tab:cost} reports the measured cost on nested subsets of 10K to
50K records. Stage~B grows from 6.2 to 66.7 seconds. Stages C and D take
18.8 and 0.4 seconds at 50K. With $K=5$, the 50,000 records
have 71,930 pairs inside clusters, below the bound $2n=100{,}000$ of
Equation~\ref{eq:vlm-calls}. Without the cap, the clusters of VL-Dedup have
457,215 such pairs, and those of HGB have 556,948,085. The VLM needs 0.437
GPU seconds per pair on one RTX~4090 with 16 parallel requests, so the pair
verdicts take about 8.7 GPU hours for all pairs of the capped clusters and
would take about 55.5 hours without the cap. The record question takes 0.066
GPU seconds per record, about 0.7 GPU hours for the 38,102 records in
clusters. The combined score decides 9,309 of the 71,930 pairs (12.9\%)
without a pair verdict, which lowers the pair verdicts to about 7.6 GPU
hours.

\begin{table}[t]
\centering
\small
\setlength{\tabcolsep}{3pt}
\caption{Stages A--D on two corpora outside VL-DedupBench. False joins and pair GPU
hours refer to the clusters of VL-Dedup. F1 is over all record pairs of COCO
and over the record pairs of the SOP test products.}
\label{tab:scale}
\begin{tabular}{@{}lrr@{}}
\toprule
 & COCO & SOP \\
\midrule
Records & 580,767 & 97,503 \\
Entities & 116,091 & 21,227 \\
Same-entity pairs & 1,162,473 & 266,739 \\
$K$ & 5 & 8 \\
Stage A (min) & 60 & 36 \\
Stages B--D (min) & 120 & 5.1 \\
\midrule
False joins, no cap & 460,000 & 15,066,652 \\
False joins, cap & 371 & 34,757 \\
F1 of VL-Dedup & 0.9989 & 0.2866 \\
F1 of B6 & 0.9866 & 0.1485 \\
F1 of B2 & 0.9970 & 0.0657 \\
\midrule
Pair GPU hours, cap & 141 & 9.2 \\
Pair GPU hours, no cap & 197 & 1,835 \\
\bottomrule
\end{tabular}
\end{table}

\paragraph{Full corpora.}
Table~\ref{tab:scale} reports Stages A--D on two corpora outside
VL-DedupBench, without calling the VLM. Stage~A runs on one RTX~4090, and
Stages B--D use eight CPU threads. The first holds every caption of the
COCO 2017 training and validation sets whose image is not in the benchmark,
with about $1.7\times10^{11}$ record pairs, and the records of one image form
one entity. We keep $K=5$, since these entities, like the COCO entities of
VL-DedupBench, have at most seven records. The second holds every SOP photo
that is in neither the benchmark nor MV, with up to 12 photos per product.
InternVL3-2B writes each caption as in Section~\ref{sec:benchmark}, which
takes 65 minutes, and validation on the bicycle and stapler products selects
$K=8$. Models, calibrators, and thresholds stay fixed. Without the cap, HGB
puts 565,417 COCO records and 51,882 SOP records into one cluster. With the
cap, VL-Dedup reaches a higher F1 than B6 and B2 on both corpora. All 371
false joins that remain on COCO come from near-identical images, whose pHash
values differ in at most 8 of 64 bits. As on MV, photos of one SOP product differ in viewpoint and background. On the test products, VL-Dedup finds 4,852 same-product
pairs, against 2,181 for B6 and 915 for B2. Stage~B returns 31.8 candidates
per record on COCO and 31.0 on SOP, against 30.1 on the 50,000 benchmark
records, so the candidates grow linearly with $n$, as the cost analysis
of VL-Dedup predicts. On both corpora, the pairs inside the capped clusters stay below the
bound of Equation~\ref{eq:vlm-calls}, and the record questions of Stage~E
would add about 11 GPU hours on COCO and 0.8 on SOP.
Without the cap, the pair verdicts on the clusters of HGB would take about
$1.9\times10^{7}$ GPU hours on COCO.

\FloatBarrier
\section{Discussion}
\label{sec:discussion}

\paragraph{Settings for a new corpus.}
VL-Dedup takes every setting from training and validation data. On the full
COCO and SOP corpora, the models, calibrators, and $\tau$ stay those of
VL-DedupBench, and on SOP only the size cap is chosen again. The validation data can be small. On the
full SOP corpus, validation on the bicycle and stapler products, 2 of the 12
SOP superclasses, selects a cap close to the best one on test. On the full COCO
corpus, the entities have the sizes of the COCO entities of VL-DedupBench, so
the cap chosen on the benchmark needs no new validation.
The threshold $t$ of the combined score also needs validation data, records
with swapped captions. Such records can be made on any corpus by moving
captions between images, as the M records of VL-DedupBench are.

\paragraph{Human review.}
The export holds a record for review instead of deleting it when the record
lies on a pair with a mismatch action or a verdict flag. In the 50 sampled
test clusters, it holds 26 of 237 records, 11\%. Besides the mismatch
actions, the VLM adds verdict flags to 3.5\% of the same-entity test pairs
inside clusters, and our audit of natural captions found wrong captions more
often in flagged pairs than in control pairs. Stage~F stores the clusters, the actions, and the
verdicts, so a reviewer sees which action held each record.

\paragraph{Checking a retained set.}
We check deletions with two AI auditors and report a range. The strict count
needs both auditors to agree, and the lenient count needs one. A human then
checks only the cases on which they disagree. In our audits of hard negatives
and of natural caption errors, these cases were few, and the human counts
fell inside the range of the two auditors. We report the lost records of each
export policy as such a range. A curator can run this check on a sample of deletions before
deleting records from a whole corpus.

\paragraph{Evaluating other systems.}
VL-DedupBench scores any system that outputs duplicate groups. The graph level
needs only the group of each record, so our experiments score
Data-Juicer, SemHash, SemDeDup, and caption MinHash from their duplicate
groups alone. A system that outputs pair decisions, such as B2 and B6, can
also be scored on the labeled pairs, and a system that outputs the records it
keeps can be scored at the retained-set level with the audit above. The three
levels thus compare systems that differ in what they output.

\paragraph{Other record types.}
The cap reads only the selected edges and their probabilities. It works with
any pair classifier that outputs a probability, and it repairs the clusters of both logistic regression and gradient boosting. ER pipelines
on text or product records turn pair decisions into clusters in the same way,
so they can bound their clusters and choose the bound on validation data.
Equation~\ref{eq:vlm-calls} then also bounds any cost per pair inside
clusters, such as a call to a large language model for matching
\citep{narayan2023llmem,peeters2025llmem}. The retained-set check needs no
images either. It compares each deleted record with the retained records of
its cluster, so it applies to any corpus in which a deletion keeps another
record of the same visual entity.

\section{Conclusion}
\label{sec:conclusion}

We present VL-Dedup, a system for graded multimodal ER on image--text
corpora. It limits the size of each visual-entity cluster, asks a VLM about
the pairs inside clusters, and scores each record to flag swapped captions.
Our results show that correct pair decisions do not ensure correct clusters,
and correct clusters do not ensure a retained set that keeps every fact, so
each level needs its own evaluation. Labeled pairs hide graph errors even when
they score clusters, because the errors spread along paths of edges and join
records that no edge compares. Results on the original records overstate
robustness, so a benchmark also needs edited copies and several photos of one
product. The same VLM judges single records better than pairs, and pairs
inside clusters better than pairs with uncertain scores. The export of
VL-Dedup loses facts in far fewer deletions than keeping one record per
cluster, because it chooses which records to delete, not only because it
deletes fewer. We plan to train vision--language
models on the retained sets of different export policies and use the
downstream results to improve the export policy. In addition, we plan to
extend graded multimodal ER and its three evaluation levels to other curation
tasks, such as ER on text records with large language models and the
deduplication of interleaved image--text documents.

\bibliographystyle{ACM-Reference-Format}
\bibliography{references}
\end{document}